\documentclass[manuscript,screen]{acmart}
\usepackage{enumerate,comment,graphicx,psfrag}
\usepackage{xspace,afterpage,lscape}
\usepackage{chngcntr}
\counterwithin{table}{section}

\usepackage[normalem]{ulem}
\usepackage{bm}
\usepackage{mathrsfs, setspace,multirow}
\usepackage{natbib,color,colortbl,url}
\usepackage[table]{xcolor}
\usepackage{caption, subcaption}
\usepackage{epstopdf}
\usepackage{breakcites,arydshln}
\usepackage[english]{babel}
\UseRawInputEncoding
\usepackage{hyperref}
\usepackage{bbding}
\usepackage{enumitem}
\usepackage[table]{xcolor}
\usepackage{adjustbox}
\usepackage{tabularx}

\definecolor{darkgreen}{rgb}{0.0, 0.5, 0.0}
\definecolor{lightgreen}{rgb}{0.5, 1.0, 0.5}
\definecolor{yellow}{rgb}{1.0, 1.0, 0.0}
\definecolor{orange}{rgb}{1.0, 0.65, 0.0}
\definecolor{red}{rgb}{1.0, 0.3, 0.3} % Reduced shade of dark red
\makeatletter
\def\ctext#1{\expandafter\@ctext\csname c@#1\endcsname}
\def\@ctext#1{\ifcase#1\or \textbf{Assumption 1}\or \textbf{Assumption 2}\or \textbf{Assumption 3}\or
\textbf{Assumption 4}\or \textbf{Assumption 5}\or \textbf{Assumption 6}\or \textbf{Assumption 7}\fi}
\AddEnumerateCounter{\ctext}{\@ctext}{Assumption 2}
\makeatother
\newlist{assumptions}{enumerate}{1}
\setlist[assumptions,1]{label={\ctext*},align=left}

\AtBeginDocument{%
  }

\setcopyright{acmlicensed}
\copyrightyear{2025}
\acmYear{2025}
\acmDOI{XXXXXXX.XXXXXXX}

\begin{document}

\title{Selecting for Less Discriminatory Algorithms: A Relational Search Framework for Navigating Fairness-Accuracy Trade-offs in Practice}

\author{Hana Samad}
\email{hsamad@nationalfairhousing.org}
\orcid{0009-0003-3111-3446}
\affiliation{
  \institution{Responsible AI Lab, National Fair Housing Alliance}
  \city{Washington}
  \state{DC}
  \country{USA}}

\author{Michael Akinwumi}
\email{makinwumi@nationalfairhousing.org}
\orcid{0009-0006-3863-0683}
\authornote{Corresponding Author}
\affiliation{
   \institution{Responsible AI Lab, National Fair Housing Alliance}
  \city{Washington}
  \state{DC}
  \country{USA}
  \orcid{0009-0006-3863-0683}
  }

\author{Jameel Khan}
\email{jkhan@nationalfairhousing.org}
  \orcid{0009-0008-5740-466X}
\affiliation{
   \institution{Housing and Community Development, National Fair Housing Alliance}
  \city{Washington}
  \state{DC}
  \country{USA}
  }

\author{Christoph M\"{u}gge-Durum}
\email{cmugge-durum@nationalfairhousing.org}
\orcid{0009-0002-8595-4112}
\affiliation{
   \institution{Responsible AI Lab, National Fair Housing Alliance}
  \city{Washington}
  \state{DC}
  \country{USA}
  }

\author{Emmanuel O. Ogundimu}
\email{emmanuel.ogundimu@durham.ac.uk}
\orcid{ 0000-0001-9252-9275}
\affiliation{
   \institution{Department of Mathematical Sciences, Durham University}
  \city{Durham}
  \state{County Durham}
  \country{UK}
  }


\begin{abstract}
As machine learning models are increasingly embedded into society through high-stakes decision-making, selecting the right algorithm for a given task, audience, and sector presents a critical challenge, particularly in the context of fairness. Traditional assessments of model fairness have often framed fairness as an objective mathematical property, treating model selection as an optimization problem under idealized informational conditions. This overlooks model multiplicity as a consideration---that multiple models can deliver similar performance while exhibiting different fairness characteristics. Legal scholars have engaged this challenge through the concept of Less Discriminatory Algorithms (LDAs), which frames model selection as a civil rights obligation. In real-world deployment, this normative challenge is bounded by constraints on fairness experimentation, e.g., regulatory standards, institutional priorities, and resource capacity. 

Against these considerations, the paper revisits \citet{lee21}'s relational fairness approach using updated 2021 Home Mortgage Disclosure Act (HMDA) data, and proposes an expansion of the scope of the LDA search process. We argue that extending the LDA search horizontally, considering fairness across model families themselves, provides a lightweight complement, or alternative, to within-model hyperparameter optimization, when operationalizing fairness in non-experimental, resource constrained settings. Fairness metrics alone offer useful, but insufficient signals to accurately evaluate candidate LDAs. Rather, by using a horizontal LDA search approach with a relational trade-off framework, we demonstrate a responsible minimum viable LDA search on real-world lending outcomes. Organizations can modify this approach to systematically compare, evaluate, and select LDAs that optimize fairness and accuracy in a sector-based contextualized manner.
\end{abstract}


\begin{CCSXML}
<ccs2012>
   <concept>
       <concept_id>10010147.10010257</concept_id>
       <concept_desc>Computing methodologies~Machine learning</concept_desc>
       <concept_significance>500</concept_significance>
       </concept>
   <concept>
       <concept_id>10010405.10010481.10010484.10011817</concept_id>
       <concept_desc>Applied computing~Multi-criterion optimization and decision-making</concept_desc>
       <concept_significance>500</concept_significance>
       </concept>
   <concept>
       <concept_id>10010405.10010455.10010458</concept_id>
       <concept_desc>Applied computing~Law</concept_desc>
       <concept_significance>500</concept_significance>
       </concept>
   <concept>
       <concept_id>10010147.10010178.10010205</concept_id>
       <concept_desc>Computing methodologies~Search methodologies</concept_desc>
       <concept_significance>500</concept_significance>
       </concept>
   <concept>
       <concept_id>10010147.10010178</concept_id>
       <concept_desc>Computing methodologies~Artificial intelligence</concept_desc>
       <concept_significance>500</concept_significance>
       </concept>
 </ccs2012>
\end{CCSXML}

\ccsdesc[500]{Computing methodologies~Machine learning}
\ccsdesc[500]{Applied computing~Multi-criterion optimization and decision-making}
\ccsdesc[500]{Applied computing~Law}
\ccsdesc[500]{Computing methodologies~Search methodologies}
\ccsdesc[500]{Computing methodologies~Artificial intelligence}

\keywords{Less Discriminatory Alternatives, LDA search, fairness-accuracy, horizontal LDA search, HMDA, financial services, sector-based, model multiplicity, relational trade-offs, operationalizing fairness}

\received{24 November 2025}

\maketitle

\section{Introduction}
As artificial intelligence (AI) and machine learning (ML) systems, hereafter referred to as algorithmic systems, are increasingly integral to housing-related decision-making, critically exploring and addressing discriminatory legacies becomes paramount. Within the housing industry, algorithmic systems have been employed to broaden access to nontraditional applicants and increase efficiency through automated underwriting systems (AUS), pricing systems, automated valuation systems (AVS), and tenant screening systems (TSS). Yet, the risk associated with these applications  remains inadequately addressed. 

Discriminatory policies and exclusionary laws fundamentally skewed generational access to financial credit and housing opportunities, thus people of color are inherently underrepresented in financial service and homeownership data. 2021 Home Mortgage Disclosure Act (HMDA) data reflects this idea, revealing significant racial disparities in homeownership rates. The rates found in these foundational datasets are the basis for algorithmic system training, informing current access and data-driven material allocation decisions. Unlike humans, who can demonstrate empathy and perhaps adjust decisions accordingly (see \citet{das2023algorithmic}), algorithms operate with less transparency and are not inherently equipped to modify or contextualize their outputs in the face of ethical considerations or sociohistorical context. We refer interested readers to learn more about this algorithmic dilemma in relation to the nuances of the housing context, as well as the associated history of discrimination in the following articles ( \citep{us1988},\citep{ffiec_hmda}, \citep{TurnerLee2018-TURDRB},  \citep{ui2020}, \citep{draude2020situated}, and Section 2 of \citep{lee21}). 

Although this study focuses primarily on mortgage lending and housing, the issues and questions explored are directly applicable to other sectors grappling with the same journeys and impediments to adopting automation. Choosing to offer AI-based services and products has serious implications for the risk management processes of private sector entities throughout the product lifecycle, including model procurement, development, and compliance to mitigate consumer harm. What liability exists in AI-based products and where opportunities for risk management exist in the development pipeline are the key questions on industry's mind. This in practice---a lack of actionable harm mitigation strategies, frameworks, and common industry standards---directly results in industry hesitance to implement algorithmic systems within federally flagged high-impact areas \citep[p.~21]{OMB2025AI}.  

Against these outstanding concerns, theoretical approaches have emerged that seek to address real-world challenges and current risk factors with AI applications. One such approach involves the identification and preference for less discriminatory algorithms (LDAs), models that demonstrate performance comparable to existing industry algorithms, while minimizing discriminatory, or disparate impact. In high-impact public facing domains such as housing, lending, or employment, where embedded data-level bias can exacerbate disparities, LDAs present a paradigm for approaching model selection to ensure business compliance with existing civil rights and other relevant legal statutes. 

In this work, we focus on an inference stage, trade-off curve analysis technique that \citet{lee21} developed and applied to 2011 HMDA data. The methodology was framed as a balance of objectives rather than an absolute mathematical fairness criterion; we apply this technique to more recently available 2021 HMDA data, ten years later. This longitudinal approach aims to assess the robustness of the relational fairness tradeoff approach across current patterns in mortgage data in America, testing if the methodology: 1) can address the contextual complexity of bias and fairness in this area, and 2) function as a clear, low barrier framework that public and private decision-makers can employ to select fairer algorithmic systems. 

This study adds new insights on three key issues by introducing the framework to ongoing legal, ML, and algorithmic governance discussions, namely the utility of model multiplicity and LDAs in practice. First, we contribute to the growing LDA literature, particularly existing work on within-model hyperparameter optimization approaches (\citep{gillis2024}; \citep{laufer2025})---what we refer to as vertical LDA searches---by arguing that the scope of the LDA search itself should be expanded to include testing model families themselves initially as a key fairness consideration. This process we refer to as a horizontal LDA search---an early-stage model development intervention that exploits structure-level algorithmic differences for improved sector-dependent fairness gains. Institutions can reproduce this internally within their existing development processes to achieve fairness and compliance in a lightweight, scalable, and reasonable manner. Secondly, we explore the relational approach in navigating the intersection of financial inclusion and racial equity within mortgage lending practices. We highlight the effects of including or excluding race as a feature in algorithmic decision-making in lending, evaluating how this impacts real-world material outcomes and predictive performance, in a sector where the foundational data is fundamentally compromised. Finally, we take these ideas in tandem and translate them into recommendations that industry and regulatory actors can focus on to successfully encourage the broader adoption of LDAs as a paradigm for legible, compliance-driven, and socially responsible AI development.

The remainder of the paper is organized as follows to build upon our central thesis. Section \ref{sec:literature review} lays the theoretical foundations for our argument, details the relational trade-off methodology, algorithms, metrics used, and reviews model multiplicity and LDAs. Section \ref{sec:methods} documents methods and alterations to the original framework's assumptions. Section \ref{sec:results} presents our empirical findings. Finally, Section \ref{sec:discussion} grounds the results and methodology employed in this study to practical application, guiding public and private entities to understand the relevance for their own interests and processes. We offer a survey of appropriate legal statutes and position this fairness framework as an evaluation mechanism for model selection. Finally, we offer prospective avenues for future work with a view towards broadening at scale adoption of responsible LDA search and turning theory to operationalization.

\section{Literature Review \label{sec:literature review}}

Fairness is not a singular or objective concept, rather it is a contested space where competing domain-contextual definitions, mathematical formulas, and normative values intersect. Addressing how to achieve fairness as a standard for algorithmic decision-making systems requires not only an understanding of fairness in theory, but a further examination of how it is measured and operationalized in practice. While frameworks exist, their successful adoption into development workflows has often been restricted by the perception of having fairness and accuracy at odds, theoretically infinite model options should an LDA search be undertaken, and the challenge of identifying an LDA  in light of competing priorities. 

In trying to understand and evaluate these separate methodological tensions, this literature review can be broken down into the following parts: 1) Understanding the conceptual landscape of fairness in machine learning, 2) Discussing fairness metrics, their use cases for selecting fairer models, 3) Evaluating the relational approach to fairness-accuracy trade-offs as the basis for a potential framework in real-world cases, and 4) Examining the applicability of model multiplicity and scope of LDAs.

Through these themes, this review seeks to clarify the challenges of practical implementation of an LDA search that find fairness forward models, and whether existing frameworks can provide viable pathways towards identifying models that mitigate bias without detriment to predictive performance.

\subsection{Defining Fairness for Algorithmic Decision-making}

The increasing deployment of algorithmic decision-making systems across various sectors has brought concerns about fairness and potential discrimination to the forefront. Particularly in sensitive areas like mortgage lending, there is a risk that algorithms, often trained on historically biased data, may perpetuate or even amplify existing societal inequalities \citep{lein}. In response, the field of algorithmic fairness has burgeoned, proposing numerous mathematical formalizations intended to capture different notions of fairness. However, many of these definitions are presented as absolute, universally applicable conditions, potentially overlooking the intricate realities of specific applications. 

\citet{lee21} challenge this paradigm, arguing that the practical and ethical trade-offs inherent in complex decision-making processes are unavoidable. They propose a shift towards understanding fairness not as a static, absolute mathematical state to be achieved, but as a relational notion. This perspective advocates for evaluating the fairness of an algorithmic system relative to other viable options and explicitly considering the trade-offs between various, often competing, objectives. Applying this to 2011 Home Mortgage Disclosure Act (HMDA) data, they emphasize the limitations of standard mathematical metrics in capturing the nuanced ethical dilemmas faced by lenders at face value. This is particularly important for lending given biases embedded within legacy data, and the difficulty of untangling legitimate risk proxies from potentially discriminatory proxies for protected characteristics, e.g., race. The relational approach encourages an evaluation focused on achieving a model as close to equilibrium between desired outcomes, such as enhancing financial access, and potential negative impacts on specific groups. It marks a significant conceptual move, steering the literature away from the pursuit of a singular, perfect mathematical fairness definition towards a more pragmatic, context-dependent framework centered on navigating and managing competing values.

Algorithmic fairness metrics are broadly categorized under two main paradigms: individual fairness and group fairness \citep{anderson2025algorithmic}. Individual fairness (IF) adheres to the principle that "similar individuals should be treated similarly". In this context, similarity is determined based on task-relevant characteristics, independent of protected attributes. For instance, two loan applicants with comparable financial profiles should ideally receive similar loan decisions. However, defining an appropriate, context-specific similarity metric remains a significant challenge, and subjective or data-driven definitions risk encoding existing biases. Group fairness (GF), conversely, focuses on achieving parity in outcomes or statistical measures across different demographic groups, typically defined by protected attributes such as race, sex, religion, national origin, or familial status \citep{gohar2023survey}. The objective is often to equalize specific rates, like approval rates or error rates, between these groups \citep{zhou2022group}.

A critical tension exists between these two paradigms: satisfying group fairness criteria does not guarantee individual fairness, and enforcing group-level parity can sometimes necessitate treating similar individuals differently based on their group membership, directly conflicting with the IF principle \citep{zhou2022group}. This conflict often reflects a deeper ethical divergence between ensuring procedural consistency for individuals (treating like cases alike) and pursuing distributive justice to rectify group-level disparities, which may be rooted in historical bias reflected in the data. Furthermore, focusing solely on isolated protected attributes can obscure intersectional discrimination experienced by individuals with multiple identities (e.g., race and gender) \citep{gohar2023survey}.

To address identified biases and strive towards fairness goals, various intervention techniques have been developed, typically categorized according to the stage of the machine learning pipeline at which they are applied: pre-processing, in-processing, and post-processing \citep{dieng2025algorithmic}.

\begin{enumerate}
    \item Pre-processing techniques: These techniques involve modifying the training data before a model is trained, removing or mitigating biases present in the data itself. Examples include suppressing sensitive attributes or their proxies, re-labeling historically biased outcomes, re-weighting data points to balance group representation, or learning on transformed data representations designed to be fair \citet{nielsen2020practical}. The main advantage is that these techniques are model-agnostic and tackle bias at its source; some research suggests that successful pre-processing to approximate an unbiased world could resolve fairness-accuracy trade-offs \citep{lein}. However, pre-processing can significantly alter the original data, simple suppression is often ineffective due to correlations, and relabeling can be subjective and practically challenging.

    \item In-processing techniques: They integrate fairness considerations directly into the model training process, typically by modifying the learning algorithm or its objective function \citep{lein}. Common approaches include adding fairness-related regularization terms to the loss function to penalize unfair predictions \citep{petersen2021post} or imposing fairness constraints during optimization. In-processing methods can potentially achieve a better balance between fairness and accuracy by optimizing them jointly \citep{zemel2024group}.  These in-processing methods include in-training approaches such as adversarial debiasing \citep{zhang_2018} and distribution matching \citep{QuadriantoSharmanska2017} techniques. The latter of which was applied to mortgage underwriting and pricing models in a study which found that distribution matching reduced racial disparities in loan approvals by 13\% and decreased borrowing costs for Black and Hispanic applicants by 20\% without compromising system accuracy \citep{merrill2024fairness}. These methods promote fairness while maintaining efficiency, however, drawbacks include often being model-specific, requiring access to and modification of the training pipeline, and incurring the computational cost of retraining \citep{zemel2024group}.

    \item Post-processing techniques: They adjust the outputs of an already trained model without altering the model itself \citep{zemel2024group}. Examples include setting different classification thresholds for different groups or learning a separate transformation function for the model's scores. The primary appeal lies in being model-agnostic, avoiding costly retraining, and applicability to deployed or black-box systems. However, their effectiveness is constrained by the information contained in the original model's predictions, they may be less impactful than in-processing methods, and can sometimes improve group fairness at the expense of individual fairness or accuracy.
\end{enumerate}

The choice among these intervention strategies is not merely technical but involves navigating practical constraints---such as access to data, control over the model training process, and computational resources---alongside strategic goals regarding the desired fairness definition and tolerance for accuracy trade-offs \citep{lein}.

\subsection{Quantifying Fairness: A Review of Mathematical Fairness Metrics}

In conjunction with theoretical fairness discussions, the field has created a toolkit of fairness metrics to quantify and describe variability between different subgroups within a dataset----in the housing sector this usually includes demographic parity, standardized mean deviation, disparate/adverse impact, and mean absolute deviation. A common definition for fairness in this context points to opportunity access rather than to outcomes themselves \citet{lee21}. Equalization of evaluation metrics attempt to formalize this belief that a model's predictive accuracy should not vary between racial groups. The most utilized and cited of these include equal opportunity (EOP), false positive error rate balance, equal odds, positive predictive parity, positive class balance, and negative class balance.  

Equal opportunity, as discussed in \citet{Hardt2016} ensures equal chances of loan or housing approval for all credit-worthy applicants, focusing on minimizing false negative rates across demographics. \citet{Chouldechova2017}'s work on false positive error rate balance seeks to ensure that similar non-credit worthy applicants are equally likely to be denied loans, preventing bias in loan disapproval. Equal odds requires algorithms to achieve equal true positive and false positive rates across groups, ensuring accuracy in identifying credit-worthy and non-worthy applicants, irrespective of demographic background \citet{Hardt2016}. Positive predictive parity, aims for equal probabilities of correct loan repayment predictions across different groups, ensures that the algorithm's trust is uniformly distributed, preventing skewed good faith predictions that favor specific demographics. Finally, \citet{Kleinberg2016}'s work on positive class balance and negative class balance ensures that the confidence in predictions---whether for loan repayment or default---is fairly balanced across groups, reflecting an objective assessment of an applicant's financial behavior. 

Collectively, fairness metrics describe data variance and facilitate the evaluation of outcome distributions. For developers, policymakers, and stakeholders navigating the concerns associated with algorithmic decision-making, they act as a key benchmarking strategy for documenting how 'fairness' within an algorithm is distributed across subgroups in a dataset.  However, as remarked in \citet{lee21}, while adept at quantifying disparities, it is also essential to note that these traditional metrics solely reflect relative comparisons between groups, similar individuals, or group error rates, without necessarily accounting for the biases embedded within the data inherently. This can result then in different effect distributions if comparing the conclusions suggested by metrics with real-world consequences. This is demonstrated in   \citet{noriega2020algorithmic}'s applied case study in algorithmic targeting of social policy in Columbia and Costa Rica. While finding overwhelming positive predictive improvements by using AI-based systems compared to the status quo, their analysis underscores that the degree of these improvements were not evenly dispersed. Error reduction ranged between 5.74\% to 22.57\% in some cases, across various urban-rural, geographic, gender, and familial divisions, highlighting the importance of documenting and testing fairness outcomes across segmented subgroups in high-impact use cases. Fairness metrics, though helpful in highlighting disparities, often present challenges in implementation, particularly in cases when the proxies for outcomes are closely intertwined with those for race, meaning they are best used when heavily contextualized to the data's use case. 

\subsection{Critical Evaluation and Adaptation of \citet{lee21}: Strengths and Limitations of the Relational Trade-off Methodology}

While informed by the work of \citet{lee21}, the research presented herein adapts certain assumptions employed in their analysis, based on limitations identified within their methodology and the specific requirements of this study. The methodology proposed in Section 5 of the paper offers a distinct approach centered on relational trade-offs. It involves several key steps: first, operationalizing relevant real-world objectives (e.g., defining and quantifying financial inclusion and negative impact on minority borrowers); second, building various predictive algorithms and computing metrics related to these objectives; third, identifying potential proxies for protected characteristics within the model features; and finally, selecting an algorithm by visualizing and evaluating the trade-offs between the competing objectives.

This approach possesses notable strengths. It promotes a holistic and contextual view of fairness, moving beyond abstract mathematical conditions to evaluate algorithms within their specific application context and considering multiple, potentially conflicting, objectives. A significant advantage is its transparency regarding ethical trade-offs; it makes the inherent conflicts between values (like profit versus fairness) explicit and quantifiable, compelling decision-makers to confront and prioritize these trade-offs, aligning with broader calls for tools to evaluate AI trustworthiness \citep{petersen2023risky}. This focus on measurable impacts and comparing concrete algorithmic alternatives lends the methodology practical relevance and actionability for stakeholders, in this case lenders, aiding them in selecting algorithms that align with their ethical commitments and level of risk tolerance. Furthermore, the methodology explicitly incorporates steps to identify and analyze potential proxies for protected attributes, fostering a more profound understanding of how bias might operate, even when sensitive attributes are formally excluded.

The methodology, however, also presents limitations. Its outcomes are highly dependent on the initial assumptions made when operationalizing objectives; how concepts like "financial inclusion" are defined and measured significantly shapes the resulting analysis. This leads to potential subjectivity, as the interpretation of the trade-offs and the final algorithm choice rely heavily on the decision-maker's values, potentially lacking the perceived objectivity of purely mathematical metrics. The analysis is also heavily dependent on data quality and availability; \citet{lee21} themselves noted limitations in the public HMDA data used in their case study, which lacked crucial variables like default outcomes and credit scores, impacting the robustness of the final trade-off evaluation. Moreover, quantifying all relevant factors (e.g., regulatory compliance, long-term societal effects, model interpretability) can be challenging, and the original study explored a limited scope of algorithms and stakeholder perspectives. Ultimately, the framework functions less as a definitive solution to fairness problems and more as a structured process for deliberation. It requires stakeholders to make explicit value judgments at multiple stages---defining objectives, assessing proxies, prioritizing trade-offs---thereby facilitating, rather than automating, ethical decision-making. Its strength lies in clarifying these choices, while its weakness is its reliance on the quality of the inputs and the deliberation itself.

Despite these limitations, the core methodology outlined in Section 5 is adopted for the analysis within this paper. It introduces the concept of relational evaluation of fairness, acknowledging and operating under the inherent complexities, practical constraints, and ethical dilemmas prevalent in this domain. This approach allows for a more nuanced evaluation that considers crucial factors like financial inclusion alongside the potential negative impacts on underserved groups, moving beyond simplistic metrics. Despite stated limitations, the methodology's strength lies in its more holistic view of fairness, ability to capture the ethical trade-offs that decision-makers are regularly confronted with, and its encouragement of explicit prioritization of identified values and objectives. For a closer review of the algorithms employed, see Appendix \ref{sec:models}.

\subsection{Model Multiplicity and Less Discriminatory Algorithms (LDAs)
}
Given the increasing reliance on AI systems in decision-making processes, it is imperative to carefully select models that do not disproportionately impact marginalized segments of society. The principle of model multiplicity suggests that prediction models may share a similar predictive performance (accuracy), but may differ in other aspects such as fairness, explainability, or robustness. Model multiplicity can be a result of either procedural multiplicity or predictive multiplicity \citep{dai2025}.

\textit{Procedural multiplicity} describes how models with different internal processes may come to equally accurate predictions \citep{dai2025}. Their internal differences may lie, in part, in the weight they give to different inputs or the variables they use altogether. Interestingly, they can reach the same outcome. For instance, a random forest and a linear model might achieve comparable levels of accuracy on a given task yet arrive at their predictions through fundamentally different reasoning processes \citep{black2022}. \citet{anders2020} illustrate this point through a study of credit-scoring models: although two models produced identical predictions across all data points, their underlying justifications diverged significantly, one model's decisions were influenced by gender, while the other relied on variables such as income and tax history.
   
\textit{Predictive multiplicity} describes the phenomena in which different models, despite achieving the same overall accuracy on a given task, produce conflicting predictions for specific inputs \citep{black2022}. \citet{marx2020} define predictive multiplicity as “the ability of a prediction problem to admit competing models with conflicting predictions”. In other words, two models might perform equally well in aggregate but still assign different (or even contradicting) labels to the same data point. This phenomenon is assessed using the labeled datasets accessible to model developers, typically the training or test sets. As developers only have access to a finite amount of labeled data, they cannot fully observe all potential points of disagreement between models. Instead, they must rely on estimates drawn from the data at hand. 

\subsubsection{\textbf{Model Multiplicity and Fairness} 
}
Two prominent themes have emerged in the literature on model multiplicity, one focusing on the opportunities for selecting algorithms that confer a greater fairness with little or no cost to accuracy, the other raises concerns about the apparent arbitrariness of prediction \citep{dai2025}.

Contrary to the fixed model assumption, the phenomenon of multiplicity implies that users can select models which are best aligned with their values such as fairness or interpretability without giving up accurate predictions \citep{black2022}.

\citet{sokol2024}, while applying large real-life datasets, demonstrates the prevalence of unfairness across models with equal accuracy, and the difficulty to mitigate the effect to grant each individual the most favorable outcome. Black argues that fairness (individual and group) in model predictions is not a necessary trade-off for accuracy in model selection. Building on this idea, \citet{black2022} proposes that model multiplicity can be interpreted in a way that aligns with the disparate impact doctrine specifically, by encouraging organizations to actively explore a range of equally accurate models in search of those that reduce discriminatory outcomes. These alternative models would need to match the performance of a baseline model already considered suitable for deployment. 

\citet{creel2021} touch on the troubling consequences of predictive multiplicity, particularly its potential to introduce arbitrariness into algorithmic decision-making. However, they ultimately contend that this arbitrariness becomes a serious concern only in scenarios where an \textit{algorithmic monoculture} exists---where the same type of model or system is used so pervasively that individuals are systematically denied opportunities across multiple domains. 

This suggests that unless model developers clearly define and optimize specific behaviors like fairness, robustness, or interpretability, there is little reason to expect a model will naturally demonstrate those qualities. This challenge, referred to as the \textit{problem of under specification} by \citet{damour2020}, highlights the importance of explicitly articulating the properties we want in our models. Without doing so, those desired characteristics are unlikely to emerge on their own. \citet{dai2025} propose that selecting for accuracy alone will lead to a much less fair outcome, and recommend explicitly optimizing for fairness when selecting for an algorithm. 

Practically, the theory of model multiplicity raises several questions, particularly regarding fairness and justifiability. While model multiplicity expands the range of options available during model selection, it may simultaneously place responsibility on developers to use that flexibility thoughtfully and to provide clear justifications for their chosen approaches \citep{black2022}. This responsibility has been referred to as the obligation to search for less discriminatory algorithms (LDAs), outlined in the following section. 

\subsubsection{\textbf{Less Discriminatory Algorithms }
}
In recent years, the concept of LDAs has gained traction at the intersection of law, ML fairness, and compliance. LDAs are, as the name suggests, algorithms that can achieve a specified task with notably less disproportionate harm or negative consequences to individuals in marginalized groups. A growing body of literature has explored the LDA paradigm which \citet{Black2023Algorithms} proposed, particularly within high-impact algorithmic contexts, as well as the technical and practical constraints of defining and operationalizing such searches. 

\citet{black2024} and \citet{gillis2024} lay the legal and normative foundations for LDA obligations, with \citet{black2024} cleanly articulating a case for a duty to search for LDAs as a necessary component of algorithmic accountability to avoid liability under the existing anti-discrimination legal regime. \citet{gillis2024} extends this concept through addressing the challenge of operationalizing such a search, arguing for practical tools and guidelines that help organizations move from a legal requirement to technical implementation. Their approach focuses on treating the search as an optimization problem, both identifying linear classification models which minimize discrimination given a set of business constraints, and return cases where this was not possible within the originally stated criteria, offering conclusive results either way. 

\citet{black2023} previously raised the idea of pipeline-aware fairness research, urging scholars to develop concrete tools for evaluating discrimination across end-to-end model pipelines rather than focusing solely on a model's outputs. Building on this direction, \citet{laufer2025} continues to raise critical questions around what qualifies as a "less discriminatory algorithm", arguing that this concept must be conceptualized in terms of competing objectives such as utility, risk tolerance, and business constraints. \citet{laufer2025} also directly addresses the inherent limits of LDA search, noting that the space of less discriminatory models is vast and often underspecified, and proposes ways to bound this space through reasonable temporal and resource constraints, optimization principles, and fairness constraints. Helpfully, they demonstrate that while businesses and consumers may have different utility priorities, there is room for consideration of both if included as relevant variables from the outset of model development. 

Across these diverse literatures fairness work has greatly advanced, seen through a recent spur of measurement and evaluation methods. The legal field has articulated the normative obligation to identify LDAs, but there is still significant work to be done to convert this into practice in ways that are transparent, replicable, and lightweight. Existing LDA search evaluation methods, by and large, test for fairness extensively within a single model family, relying on technically demanding processes including complex hyperparameter optimization routines, which may require non-insignificant computational resources, or technical expertise. We refer to this approach in this paper as a 'vertical LDA search'. Although rigorous and useful, relying exclusively on this approach poses practical challenges for real-world implementation, particularly in resource-constrained environments and reporting contexts, which may pose barriers to industry-scale adoption of LDAs. Our contribution seeks to bridge these gaps to encourage LDA usage as widely as possible, building on the research agenda outlined by \citet{black2023}, particularly the urgent need for operational tools and methods in high-impact industries. 

This paper seeks to situate model multiplicity as a strategic advantage in the context of LDA searches. The literature, to date, has referred to LDA search as a within-model optimization process, thus operating under the assumption that meaningful fairness variance will occur in a vertical search context. While this approach has yielded important insights, we argue that this definition only reflects one dimension of the search space---practitioners also can look for fairness gains not only within a model, but also through the initial act of model family selection itself. This mirrors similar early-stage development workflows that involve searching for preferential model families in the context of performance and accuracy goals. We introduce this cross-model fairness evaluation approach as 'horizontal LDA search'; using the relational fairness framework previously mentioned, we position it as an initial resource-efficient and scalable step to guiding LDA selection. This secondary dimension is then in complement to existing vertical search approaches, broadening the search space into a two-dimensional space. 
 
To demonstrate the utility of horizontal LDA search, we use mortgage lending as our case study. Through evaluating a variety of commonly used model families (e.g., logistic regression, KNN, CART, NB, and random forest) using standard fairness metrics, then contextualizing these metrics against their deployment context and prioritized objectives, we demonstrate that meaningful fairness tradeoffs can be visualized across candidate models, and then subsequently used to select the model family that best aligns with the specific fairness, performance, and consumer risk priorities of a given sector and organization. Through this work, we hope to ground the LDA operationalization discussion, ensuring that rigorous theory can effectively translate into commonsense approaches for the reasonable selection of an LDA within resource-constrained and complex multi-priority environments. This paper anchors itself within the emerging literature on LDAs, as a complement to existing work on vertical LDA searches, while inviting horizontal broadening of the LDA search scope to include initial evaluations of structural model design as it relates to the fairness-accuracy question.

\textbf{
\section{Methods and Methodology }
\label{sec:methods}}

This section outlines the empirical design and methodological approach used to explore the fairness-performance trade-offs we chose across ML model families in mortgage lending. We begin by detailing the primary dataset used within this study, the 2021 Home Mortgage Disclosure Act (HMDA) data, published by the \citet{ffiec_hmda_2021}, which extensively documents records on mortgage lending practices in the United States,  providing the foundation for analysis. The approach to model development for our five algorithms, including our target variable and relevant feature encoding, is disclosed. Finally, we outline the methodological assumptions underpinning our analysis, including the choices made to simulate real-world lending conditions, which attempt to balance clarity and reproducibility. 

\subsection{Case Study: HMDA 2021 Dataset}

The HMDA dataset categorizes race and ethnicity directly using the \texttt{applicant\_race\_1} and \texttt{applicant\_ethnicity\_1} fields, as shown in Table \ref{tab_Race_Eth}. For data exploration purposes, outcomes for Asian, Black, Hispanic, Native American and Pacific Islander, and Non-Hispanic White groups are displayed in \ref{sec:data visualizations} , however for graphs included in\ref{sec:model accuract section}, only Black and Non-Hispanic White data are used. 

\begin{table}[htbp]
 \centering
 \tiny % Adjust the size as needed
 \caption{Data Fields (Race and Ethnicity) with Values and Definitions}
  \label{tab:example1}  \begin{tabular}{|p{4cm}|p{4cm}|}
     \hline
     \multicolumn{1}{|c|}{\textbf{Race}} & \multicolumn{1}{c|}{\textbf{Ethnicity}} \\
     \hline
     Description: Race of the applicant or borrower & Description: Ethnicity of the applicant or borrower \\
     \rule{0pt}{1.5\normalbaselineskip}
     Values - Definitions: & Values - Definitions: \\
     1 - American Indian or Alaska Native & 1 - Hispanic or Latino \\
     2 - Asian & 11 - Mexican \\
     21 - Asian Indian & 12 - Puerto Rican \\
     22 - Chinese & 13 - Cuban \\
     23 - Filipino & 14 - Other Hispanic or Latino \\
     24 - Japanese & 4 - Not applicable \\
     25 - Korean & \\
     26 - Vietnamese & \\
     27 - Other Asian & \\
     3 - Black or African American & \\
     4 - Native Hawaiian or Other Pacific Islander & \\
     41 - Native Hawaiian & \\
     42 - Guamanian or Chamorro & \\
     43 - Samoan & \\
     44 - Other Pacific Islander & \\
    5 - White & \\
     6 - Information not provided by applicant in mail, internet, or telephone application & \\
     7 - Not applicable & \\
     \hline
   \end{tabular}
   \label{tab_Race_Eth}
 \end{table}

For this study, the categorization of race is as follows: 

\begin{itemize}    
    \item For applicants with \texttt{applicant\_race\_1} values of 2, 21, 22, 23, 24, 25, 26, or 27, and \texttt{applicant\_ethnicity\_1} values of 2, 3, or 4, the category is \texttt{Asian}.
    \item If \texttt{applicant\_race\_1} is 3 and \texttt{applicant\_ethnicity\_1} is 2, 3, or 4, the applicant is labeled as \texttt{Black}.
    \item Similarly, a \texttt{applicant\_race\_1} value of 5 with \texttt{applicant\_ethnicity\_1} values of 2, 3, or 4 results in the \texttt{White}.
    \item If \texttt{applicant\_race\_1} is 1, 4, 41, 42, 43, or 44, and \texttt{applicant\_ethnicity\_1} is 2, 3, or 4, the applicant is classified as \texttt{Other\_Race}.
    \item When \texttt{applicant\_race\_1} is 6 and \texttt{applicant\_ethnicity\_1} is 2, 3, or 4, the category is \texttt{Race\_not\_reported}.
    \item Additionally, if \texttt{applicant\_ethnicity\_1} is 1, 11, 12, 13, or 14, the applicant is categorized as \texttt{Hispanic}, regardless of their race.
    \item If none of these conditions are met, the category is labeled as \texttt{None}.
\end{itemize}

\subsubsection{Data Visualizations} \label{sec:data visualizations}

\begin{figure}[htbp]
  \centering
  \includegraphics[width=0.7\textwidth]{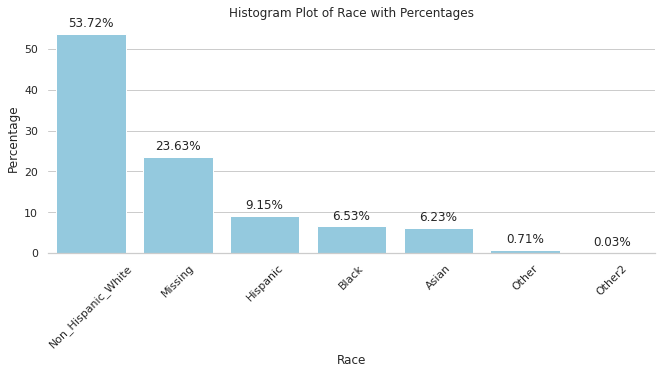}  
  \caption{HMDA 2021 Home Ownership Race Distribution}
   \label{fig_Race_Perct}
  \label{fig:sample}
\end{figure}

Figure \ref{fig_Race_Perct}  illustrates the homeownership race distribution from the employed 2021 HMDA data. It shows that White individuals make up the largest portion of homeowners at 53.72\%, followed by missing data at 23.63\%. Hispanic, Black and Asian homeowners account for 9.15\%, 6.53\% and 6.23\% respectively, while the "Other" racial category and those identifying as "2 or more" races make up smaller shares at 0.71\% and 0.03\%. This data overall highlights the distribution  of homeownership.

Figure \ref{fig_Race_Loan_ApprovedDenied} shows the distribution of loan approvals and denials in the data set within different races, with significantly higher in-group denial rates for applicants identifying as Black or Hispanic. Figure \ref{fig_AvgLoanAmount} and Figure \ref{fig_AvgIntRate} highlight the average loan amounts and average interest rates amongst racial groups.

\subsection{Model Design and Development
}
Table \ref{tab_TargetLoan} outlines the target variable used in our machine learning model, indicating whether a loan application was approved or denied based on various loan actions. It summarizes the relationship between different loan actions and the resulting approval or denial status, serving as a reference for building predictive models in the loan approval process. In this case, the positive label for prediction is "Loan Denied," and the negative label is "Loan Approved." For further information on features used to develop the model, reference Table \ref{tab_ModelFeatures}.  The features provide valuable insight into the profiles of loan applicants and their neighborhoods, which may contribute to the likelihood of loan approval or denial. For data sampling, this paper used the same data sampling technique as stated in \citet{lee21}. 
 \begin{table}[htbp]
   \centering
  \scriptsize  
   \caption{Mapping of Loan Action taken to Approved and Denied}
   \label{tab:example2}
   \begin{tabular}{|p{7.5cm}|p{3cm}|p{3cm}|p{3cm}|}
     \hline
     \multicolumn{1}{|c|}{\textbf{Action Taken}} & \multicolumn{1}{c|}{\textbf{Loan Approved or Denied}}\\
     \hline								
 	Loan Originated, Application approved but not accepted, Purchased loan, Preapproval request approved but not accepted & Approved\\
 	\hline	
     Application denied, Preapproval request denied & Denied\\	
    \hline	
   \end{tabular}
   \label{tab_TargetLoan}
 \end{table}

\subsection{Methodological Assumptions}

The foundation of this paper's analysis compares the ability of the models outlined in Section 2.4 to advance financial inclusion and minimize adverse impact on protected groups. 

The assumptions used to estimate the impact of a model on financial inclusion and protected groups are as discussed in \citet{lee21}, with the addition of Assumption 7:

\begin{assumptions}

    \item Data Representativeness:  HMDA data are assumed to reflect loan outcomes of a perfectly informed market that relies only on a single loan approval model. Under this assumption, all approved loans would get repaid and all denied loans ultimately defaulted. This aligns with the regulatory framework established by laws like the Fair Credit Reporting Act (FCRA), which requires data used in credit decisions to meet standards of maximum possible accuracy. Thus, the data used is assumed to reflect and inform accurate underwriting practices \citep{cfpb2022accuracy}.

    \item Single Lender Simplification: For analysis, the  lending market is modeled as a single entity with a maximum \$1 billion loan cap. This assumption abstracts the complexities of market dynamics into a singular analytical focus.

    \item Algorithmic Loan Decisions: Loan issuance and amounts are guided by the following process:
    \begin{enumerate}
        \item The lender uses one of the models discussed in section \ref{sec:models} one step at a time to predict whether the loan will default, i.e., whether or not the mortgage loan application should be approved or denied as discussed in Table \ref{tab_TargetLoan}.
        \item The lender sorts the loans in the decreasing order of the probability of default or increasing probability of repayment.
        \item The lender prioritizes loan approvals based on the highest likelihood of repayment until reaching the lending threshold of \$1 billion. In other words, the lender approves loans maximizing risk-adjusted returns until the \$1 billion cap is reached.
    \end{enumerate}

    \item Binary Loan Outcomes: Loans will either default or be fully repaid or defaulted. This binary treatment of loan outcomes into full repayment or default streamlines economic analyses of expected returns.

    \item Objective is to Maximize Expected Returns: The lender seeks to maximize the expected value of the loan portfolio, accounting for algorithm accuracy and loan amount. This reflects the economic imperative to balance risk and return in financial decision-making and aligns with the lender's legal obligation to shareholders and the broader community while ensuring that responsible lending practices lead to community uplifting.
    \begin{itemize}
        \item[] Note: The expected value of the loan would be the probability-weighted sum of all possible outcomes. The formula to calculate this is:
        \begin{equation}\label{expected_value} 
            \text{EV} = (P \times R) - ((1 - P) \times L), \text{where}
        \end{equation}
        \begin{itemize}
            \item EV is the expected value of the loan,
            \item P is the probability of repayment (assumed to be the model's accuracy),
            \item R is the total repayment amount if the loan is repaid (principal + interest),
            \item L is the loss if the loan defaults (often just the principal, if no recovery is assumed), and
            \item as a result of our assumptions, the equation reduces to $\text{EV} = P \times R$.
        \end{itemize}
    \end{itemize}

    \item Uniform Loan Terms: There is no racial differentiation in loan amount, interest rate, duration, type of loan or any other conditions or terms of the loans.

    \item Racial Negative Impact: We focused on Black mortgage applicants as a benchmark for potential disparities relative to Non-Hispanic White applicants, reflecting the historical context of housing discrimination. Our metric measured the direct denial rate experienced by Black applicants (denied loans / total Black applicants). While not using Adverse Impact Ratios (AIR) (see \citet{relmancolfax_upstart_2020}), this approach facilitates transparent comparison with \citet{lee21}'s findings, requiring careful interpretation while ensuring compliance with fair lending principles.
    
\end{assumptions}

Assumptions 1 to 7 provide the basis for estimating financial inclusion with the expected value of the loans when predicted risks of default are known. The primary objective of financial inclusion is to bridge the gap between those who have access to banking and financial services, including access to mortgage loans, and those who are excluded, often due to poverty or protected characteristics such as race and color, to help reduce inequality and advance economic mobility \citep{relmancolfax_upstart_2020}. Future work will be needed to investigate the updates required to apply the assumptions to low-income mortgage applicants and high-risk applicants with and without consideration of protected characteristics.

\section{Results \label{sec:results}}

In this section, we present the findings of applying \citet{lee21}'s fairness assessment methodology to contemporary lending data from the 2021 HMDA. Similarly, our analysis evaluates the performance of multiple machine learning models across different fairness metrics, quantifying the trade-offs between accuracy and fairness through variation in algorithmic structure. Overall, our findings demonstrate the robustness of the original framework and its ability to transparently visualize algorithmic trade-offs structured under complex real-world conditions. These conditions are well illustrated through the highly regulated environment that the lending case study presents.

 \subsection{Fairness Incompatibility and Trade-off Analysis}

Racial disparity measures such as those discussed in \citet{lee21}, do not account for historical bias that is prevalent in mortgage data, although they are often used in evaluating lending decisions. For example, existing algorithmic fairness metrics either measure disparity in lending decisions at group or individual levels under absolute mathematical conditions as previously discussed in Section 2.2. Hence, we tested the compatibility of different mathematical notions of fairness (see Section 2 and Figure 2 in \citet{lee21})  to decide if trade-offs such as those suggested in studies such as the aforementioned, \citet{Kleinberg2016},  and \citet{hellman2020measuring} are necessary for the models in this work.

The comparative analysis of fairness metrics across the ML models discussed in Section 2, focuses on the disparities between Non-Hispanic White applicants and applicants identified as Black or People of Color  ((\ref{tab_Black}) and \ref{tab_PoC}). The metrics---including Equal Opportunity (EOP), False Positive Error Rate Balance (FPERB), Equal Odds (EO), Positive Predictive Parity (PPP), Positive Class Balance (PCB), and Negative Class Balance (NCB)--- reveal how model choices affect algorithmic equity in lending outcomes. These results expressed in percentages, indicate the difference when Black or POC applicant outcomes are subtracted from those for White applicants. 

Negative values generally signify that the fairness metrics for White applicants are lower than those of their Black or POC counterparts, while vice versa, positive values indicate higher metrics for White applicants in comparison to their peers. The color shading and intensity reflected in the heatmap tables below correspond to the proximity of these values to zero, along a color gradient (green, yellow, orange, red). Darker shades (e.g., red) represent values further from parity.

\begin{table}[h!]
\centering
\small % Reduce font size
\caption{Assessing Equity: A Comparative Analysis of Fairness Metrics Between Non-Hispanic White and Black Applicants. }

\label{tab_Black}
\begin{adjustbox}{max width=\textwidth}
\begin{tabularx}{\textwidth}{|>{\raggedright\arraybackslash}X|X|X|X|X|X|X|}
\hline
\rowcolor{gray!20} \textbf{Fairness Metric / Model} & \textbf{Equal Opportunity (EOP)} & \textbf{False Positive Error Rate Balance (FPERB)} & \textbf{Equal Odds (EO)} & \textbf{Positive Predictive Parity (PPP)} & \textbf{Positive Class Balance (PCB)} & \textbf{Negative Class Balance (NCB)} \\ \hline
LR   & \cellcolor{lightgreen!30} 6\%   & \cellcolor{yellow!50} 10\% & \cellcolor{lightgreen!30} 6\% & \cellcolor{orange!70} 18\% & \cellcolor{darkgreen!100} 2\% & \cellcolor{darkgreen!100} 2\% \\ \hline
KNN  & \cellcolor{darkgreen!100} 5\%   & \cellcolor{darkgreen!100} 4\%  & \cellcolor{darkgreen!100} 5\% & \cellcolor{red!90} 19\% & \cellcolor{lightgreen!50} 4\% & \cellcolor{darkgreen!100} 2\% \\ \hline
CART & \cellcolor{yellow!50} 11\%  & \cellcolor{orange!60} 17\% & \cellcolor{yellow!100} 11\% & \cellcolor{yellow!50} 14\% & \cellcolor{yellow!60} 11\% & \cellcolor{yellow!50} 17\% \\ \hline
NB  & \cellcolor{yellow!70} 14\%  & \cellcolor{lightgreen!40} 5\%  & \cellcolor{orange!70} 14\% & \cellcolor{orange!70} 18\% & \cellcolor{yellow!50} 9\% & \cellcolor{lightgreen!60} 6\% \\ \hline
RF   & \cellcolor{red!100} 24\%  & \cellcolor{red!100} 43\% & \cellcolor{red!90} 24\% & \cellcolor{darkgreen!100} 3\% & \cellcolor{red!100} 13\% & \cellcolor{red!100} 18\% \\ \hline
\end{tabularx}
\end{adjustbox}
\end{table}
\footnotetext[1]{Random Forest (RF),  Logistic Regression (LR), k-Nearest Neighbors model (KNN), Classification and Regression Tree (CART), and Gaussian Na\"ive Bayes model (NB).}

\footnotetext[2]{
We used \textbf{Equal Odds} (EO) and it is calculated by taking the average of the difference in False Positive Rates and the difference in True Positive Rates between two groups. Mathematically, it can be expressed as:
$\text{EO} = \frac{1}{2} \left( \left| FPR_{\text{group1}} - FPR_{\text{group2}} \right| + \left| TPR_{\text{group1}} - TPR_{\text{group2}} \right| \right)
$.
} 
\begin{table}[h!]
\centering
\small % Reduce font size
\caption{Assessing Equity: Disparity Measures in Fairness Metrics Between Non-Hispanic White Applicants and People of Color.}
\label{tab_PoC}
\begin{adjustbox}{max width=\textwidth}
\begin{tabularx}{\textwidth}{|>{\raggedright\arraybackslash}X|X|X|X|X|X|X|}
\hline
\rowcolor{gray!20} \textbf{Fairness Metric / Model} & \textbf{Equal Opportunity (EOP)} & \textbf{False Positive Error Rate Balance (FPERB)} & \textbf{Equal Odds (EO)} & \textbf{Positive Predictive Parity (PPP)} & \textbf{Positive Class Balance (PCB)} & \textbf{Negative Class Balance (NCB)} \\ \hline
LR   & \cellcolor{lightgreen!20} 3\%  & \cellcolor{darkgreen!100}  0\% & \cellcolor{lightgreen!20}3\% & \cellcolor{red!100} 12\% & \cellcolor{darkgreen!100} 0\% & \cellcolor{darkgreen!100} 0\% \\ \hline
KNN  & \cellcolor{lightgreen!20} 2\%  & \cellcolor{lightgreen!30} 1\%  & \cellcolor{darkgreen!100} 2\% & \cellcolor{red!100} 12\% & \cellcolor{lightgreen!40} 1\% & \cellcolor{lightgreen!40} 1\% \\ \hline
CART & \cellcolor{yellow!50} 10\% & \cellcolor{yellow!50} 9\%  & \cellcolor{orange!60} 10\% & \cellcolor{orange!60} 10\% & \cellcolor{red!100} 10\% & \cellcolor{orange!60} 9\% \\ \hline
NB  & \cellcolor{darkgreen!100} 0\%  & \cellcolor{darkgreen!100} 0\%  & \cellcolor{darkgreen!100}0\% & \cellcolor{orange!60} 10\% & \cellcolor{darkgreen!100} 0\% & \cellcolor{lightgreen!40} 1\% \\ \hline
RF   & \cellcolor{red!100} 17\% & \cellcolor{red!100} 21\% & \cellcolor{red!100} 17\% & \cellcolor{darkgreen!100} 2\% & \cellcolor{red!100} 10\% & \cellcolor{red!100} 11\% \\ \hline
\end{tabularx}
\end{adjustbox}
\end{table}

Model choice, as seen through Tables \ref{tab_Black} and \ref{tab_PoC}, has significant effects on how fairness outcomes are recorded. Notably, the logistic regression model and KNN model indicate less disparity in EOP for Non-Hispanic Whites compared to other models, while the random forest model demonstrates substantial disparity towards White applicants in false positive error rate balance. The NB model shows a 14\% higher EOP metric indicating a disparity favoring White applicants, while the random forest model exhibits a 43\% lower FPERB, disfavoring White applicants (Table \ref{tab_Black}). On average, and in contrast to other models, the KNN and logistic regression models show moderate disparities across all fairness metrics. The PPP, PCB, and NCB metrics in particular highlight ongoing challenges in achieving compatibility in fairness metrics. 

Overall, the results in both Tables \ref{tab_Black} and \ref{tab_PoC} highlight the challenge of achieving compatible and consistent mathematical fairness in algorithmic decisions, showcasing the need for trade-offs techniques that aim to serve all racial groups equitably. These results underscore the complexities in algorithmic decision-making and the necessity of adopting nuanced, well-deliberated trade-off strategies to ensure equity across racial demographics for a given context. However, without additional context on the dataset, task, or sector, it would be premature to definitively generalize conclusions about the suitability of these models for real-world application based on algorithmic performance on fairness metrics alone. Subsequent analysis focuses on model performance in their original context, revealing the distribution of both accuracy and performance, within the broader lending decision-making landscape.

\subsection{Model Accuracy under Race-Aware vs. Race-Blind Conditions} \label{sec:model accuract section}

\subsubsection{Algorithmic Accuracy Comparisons}

\begin{figure}[htbp]
  \centering
  \includegraphics[width=0.8\textwidth]{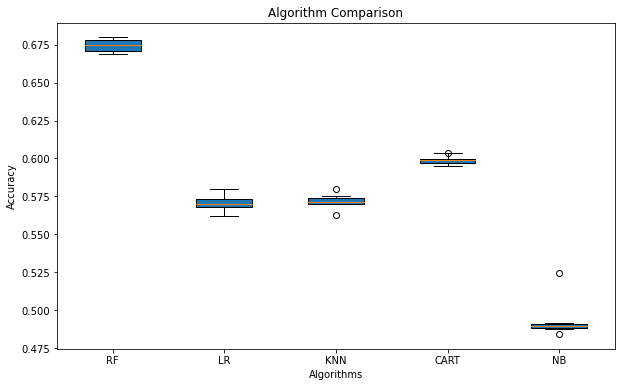}  
  \caption{Algorithm Accuracy without Demographic Features of Race, Sex, and Minority Population of the Census Tract Area}
  \label{fig_algorithm_ accuracy}
\end{figure} 

Figure \ref{fig_algorithm_ accuracy} simply presents a box plot comparison of the accuracy values for our six tested algorithms. It is important to note that the models used to generate the data for this chart did not incorporate demographic features such as race, sex, or minority population of the census tract area. The y-axis represents the accuracy values, while box plots show the distribution of these values across the different algorithms. The chart allows for a visual comparison of the performance of these algorithms, highlighting relative strengths and weaknesses in terms of accuracy.

The random forest model has the highest median accuracy above 67\%, indicating it is generally the most accurate among the algorithms presented (Figure \ref{fig_algorithm_ accuracy}). Logistic regression and KNN models also reflect relatively high and consistent accuracy. In contrast, the CART model exhibits a wider range of accuracy values, suggesting it may be more sensitive to the specific data or parameters used. The NB model demonstrates the lowest median accuracy among the five models tested.

These insights are useful for practitioners looking to select the most appropriate machine learning algorithm for their specific needs based on the required degree of performance, as seen through this lending case study.

\begin{figure}[htbp]
  \centering
  \includegraphics[width=0.8\textwidth]{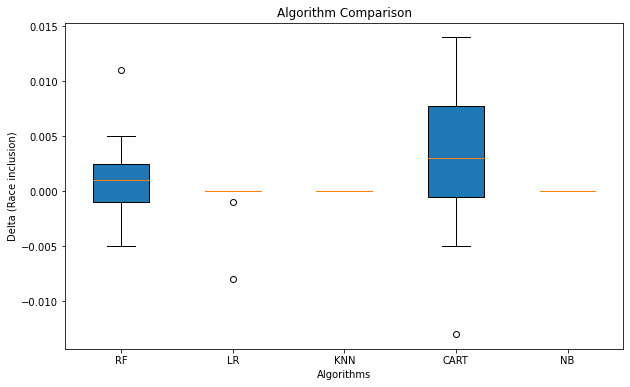}  
  \caption{Change in Algorithm Accuracy after Inclusion of Race}
  \label{fig_Change_in_algorithm_ accuracy}
\end{figure} 

Figure \ref{fig_Change_in_algorithm_ accuracy} compares the impact of including race in various machine learning models. The y-axis represents the change in the metric due to the inclusion of race in the model. Positive values suggest that including race improved fairness or performance, while negative values suggest a detrimental impact.

The random forest model demonstrates a relatively tight distribution with a slight positive median delta, suggesting that the inclusion of race marginally enhances fairness while maintaining stability. This indicates random forest's potential to improve outcomes for protected groups without introducing significant variability in performance in lending.

The logistic model exhibits minimal variation, with most of the data clustered around zero. This implies that incorporating race has little to no effect on fairness or model performance, making it a neutral choice in terms of impact. Similarly, KNN and NB also demonstrate minimal variation, making their stability comparable to the logistic regression model in terms of consistency and reliability. This suggests that including race has a neutral to minimal impact on fairness for these three models, making them consistent options for scenarios where stability is key.

In contrast to these findings, the CART model stands out for its high variability, exhibiting a wide range of values and a positive median. While this indicates the potential for significant fairness improvements through the inclusion of race, it also introduces risks of outliers where the impact could be negative. This variability highlights the model's sensitivity to fairness interventions. CART shows the highest variability, emphasizing its sensitivity to the inclusion of race, making it a powerful tool for scenarios requiring significant fairness gains, but this does necessitate careful and continual monitoring to avoid unintended consequences.

The random forest model exhibits moderate variability, making it a more stable choice while still allowing for fairness improvements. Models like logistic regression, KNN, and NB are appealing for their consistency, making them suitable for applications where stability is prioritized over significant changes in fairness metrics. In some cases, however, models like CART, with higher variability are better suited for use cases where substantial fairness improvements are desired, provided there are mechanisms in place to address potential risks. Overall, Figure  \ref{fig_Change_in_algorithm_ accuracy} underscores the critical importance of model family selection in addressing fairness issues in mortgage lending.

\begin{figure}[htbp]
  \centering
  \includegraphics[width=0.8\textwidth]{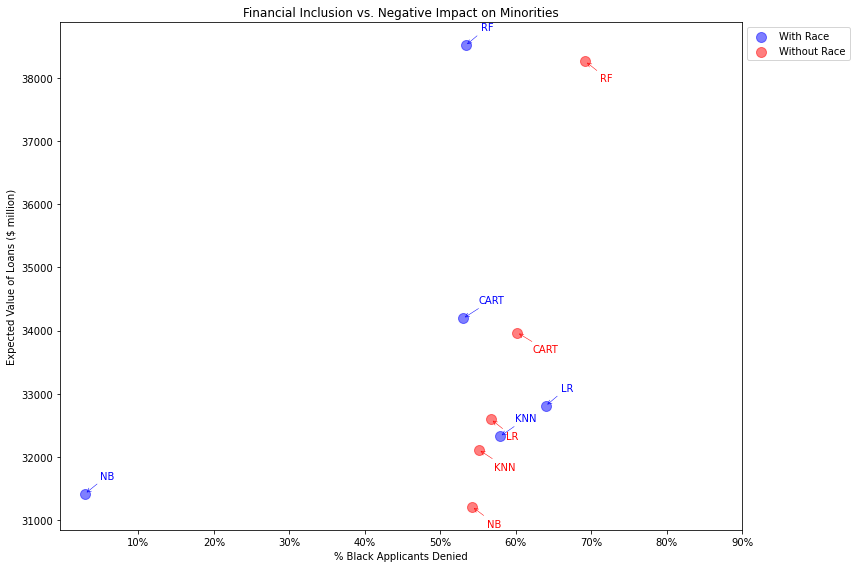}  
  \caption{Financial Inclusion vs. Negative Impact on Minorities--- with and without Race}
  \label{fig_Trade-off}
\end{figure}

Figure  \ref{fig_Trade-off} provides a comprehensive overview of how various machine learning models balance financial inclusion with minimizing negative impacts on Black applicants. The x-axis of the chart represents the percentage of Black applicants denied loans, serving as a measure of the negative impact on minorities. Meanwhile, the y-axis reflects the expected value of loans (in millions), indicating the degree of financial inclusion. The trade-off between these two objectives is clearly illustrated, emphasizing the challenge of achieving both fairness and economic returns simultaneously. Performance of models when race is explicitly included is represented by blue points, and when race excluded is represented by red points. 

Models incorporating race generally achieved higher levels of financial inclusion, reflected in higher expected loan values, and often result in lower denial rates for Black applicants. This underscores the critical role of including race as a stabilizing feature in some models for fair lending. In this context, it helps to address disparities in random forest, CART, and NB for minority borrowers. The performance of individual models, however, does vary significantly, as seen in their respective positions on the chart. The random forest model without race, for example, achieves the highest financial inclusion, but at the cost of a higher denial rate for Black applicants, highlighting a tension between maximizing economic returns and promoting equity. In contrast, NB with race demonstrates the lowest denial rate for Black applicants, prioritizing fairness over financial returns. Other models, such as CART and KNN, strike a middle ground when including race. For lenders seeking balance or consistency, these models balance financial inclusion and minimize potential risk, making them potential candidates when both objectives holding equal weight.

The analysis operates within a simplified framework defined by specific assumptions, such as a single lender, uniform loan terms, and binary loan outcomes. These assumptions create a controlled environment to evaluate model performance, but may not fully capture the complexity of real-world lending dynamics. As such, the findings should be viewed as a foundation for further investigation rather than definitive conclusions, and the approach and assumptions for other use cases outside of lending should be considered and modified appropriately.

\section{Discussion \label{sec:discussion}}

The prior analysis of fairness across machine learning models carries several implications for model developers, deployers, and users. For one, there is no single best algorithm for fairness. Secondly, existing approaches are susceptible to structural bias embedded in the input data and to proxies associated with race or other protected characteristics. The question remains, however, of how model developers should approach selecting the appropriate model for a given task.  

\citet{lee21}'s technique uses relative trade-offs between objectives as a marker for fairness. This approach emphasizes flexibility for each individual context, allowing distinct variables to be considered in relation, and applies a relative evaluation scheme by comparing different algorithms. Based on a decision-maker's specific goals and context, a different algorithm may be suggested, providing the best balance between different objectives and thus may represent the most “fair” approach for that use case. There is no single "one-size-fits-all" method that guarantees optimal fairness. Instead, decision-makers must balance objectives according to their unique circumstances, and consider what information is required to achieve fairer outcomes in their models (e.g., inclusion or exclusion of race depending on model family). A larger loan can generate higher revenue for a company, but loan rejections also carry the risk of financial loss. A higher loan size, in turn, can lead to a greater rejection rate, as seen in the comparison between models e.g., random forest (RF) and Na\"{i}ve Bayes (NB) without race. In the lending context, the trade-off between financial inclusion of minority borrowers and the accuracy of predicting loan defaults requires careful calibration, and a moderate approach may be preferred. For cases similar to the one described in this study, this may mean use of CART, random forest, or KNN models, for example.

The implications of the fairness metric analysis (Tables \ref{tab_Black} and \ref{tab_PoC}) in conjunction with accuracy tests (Figures \ref{fig_algorithm_ accuracy} and  \ref{fig_Change_in_algorithm_ accuracy}) present what a holistic look at fairness and LDA searches must consider--- that internal model design for fairness has to be considered within the specific real-world context that the automated system will be providing material outcomes in. While reflections on fairness metrics such as Equal Opportunity or Equalized Odds provide important insight into the relative treatment of applicants, our findings suggest that relying on these metrics alone may obscure the material realities that the model's decisions produce for subgroups. For example, random forest models demonstrate larger fairness disparities, indicating disadvantage for White applicants in certain metrics (e.g., FPERB;  Table\ref{tab_Black}). However, when assessing actual loan denial rates and level of financial inclusion for Black applicants, random forest models operationalized still have significantly higher denial rates relative to the other models tested (Figure \ref{fig_Trade-off}). This was especially without the inclusion of race as a feature.

The assumption that an inherent trade-off exists between reducing disparity and maintaining accuracy has also been challenged by recent computer science research. In this context, disparate impact refers to the disproportionate negative effects a policy or model may have on disadvantaged groups, while accuracy reflects the model's effectiveness in achieving its predictive goals, such as assessing loan default risk. Model multiplicity demonstrates that multiple models can achieve similar predictive accuracy while exhibiting vastly different levels of disparate impact \citep{black2022}. This means that companies are not necessarily forced to choose between fairness,  performance, and accuracy---they can be optimized in conjunction if properly contextualized fairness considerations are integrated into model development from the outset. Thus, rather than settling for models that perpetuate bias, institutions should actively search for less discriminatory algorithms that maintain predictive accuracy while reducing disparities. 

Given that no single "silver bullet" algorithm that can ensure fairness similarly across all scenarios, institutions must move towards adopting a strategic and dynamic approach towards algorithmic selection, based on their sector's particular history and available data, to maintain strong predictive capabilities. This necessitates evaluating and refining models to prioritize fairness, and understanding what effect the inclusion or exclusion of features does to model performance across various families, to help or to harm different subgroups. By leveraging model multiplicity from the outset, companies can proactively seek models that align with both business objectives and ethical responsibilities. 

\subsection{Future Directions and Recommendations}

The financial benefits of adopting AI in the financial services industry are well-documented, including increased efficiency and broadening credit, mortgage, and financial access to traditionally excluded customers.  A recent US \citet{schaffer2024} survey report, however, highlighted that lenders and insurers are still reluctant to adopt AI in their decision-making processes. The responsibility associated with creating and deploying public-facing automated decision-making tools for resource allocation or opportunity access is a heavy one for public and private sector entities to hold. AI is a technology that in many ways is still in its infancy, and developing sensible risk management infrastructure in model procurement, development, and deployment processes are outstanding considerations before mass adoption.  

These concerns ultimately revolve around questions of liability—where to assign legal responsibility for the decisions of automated systems, and how to successfully and precisely mitigate unintended harms before they occur. The current legal regime applicable to automated systems consists of existing civil rights statutes such as Title VII, the Equal Credit Opportunity Act (ECOA), and the Fair Housing Act. Also directly applicable have been interpretations of algorithmic harm through the lens of disparate impact theory, as seen in Title VI (\citep{black2024}). In the past two years, LDAs in particular have received attention from AI practitioners and federal bodies as a paradigm to achieve compliance with existing civil rights law. CFPB's annual Fair Lending report \citep[p.36]{cfpb2024} made what responsibilities that entails clear for lenders and insurers, "Robust fair lending testing of models should include regular testing for disparate treatment and disparate impact, including searches for and implementation of less discriminatory algorithms using manual or automated techniques.''

This study has made the case for one high-impact evaluation framework in the statistical modeling stage of AI model development, a re-contextualized fairness framework that is capable of being easily integrated into the existing algorithmic development process pre-deployment for public interacting models. Previous corporate attempts (e.g., Upstart) to include fairness interventions with LDAs have highlighted the critical nature of not only having solutions that are impactful in improving outcomes across protected classes, but making sure they are time efficient, cost-effective, and accessible to organizations of all sizes \citep{colfax2024}. Conducting a horizontal LDA search is one such lightweight fairness intervention that can be taken to guide LDA searches for resource-constrained institutions, or as a first step before more computational intensive approaches---such as vertical LDA searches---for organizations wanting to commit to a more comprehensive search process. For relevant policy and development recommendations, see Appendix \ref{policy recommendations}

\section{Conclusion}

This paper has demonstrated the complexity of balancing fairness and accuracy in automated decision-making, particularly within the heavily regulated financial industry, where the stakes are high and the space for experimentation is narrow. Although no single algorithm guarantees optimal fairness, continuing to leverage the concept of model multiplicity offers a promising path forward, demonstrating that multiple model families can achieve comparable predictive accuracy, while exhibiting varying levels of disparate impact. Formalizing an LDA search as an institutional or regulatory convention requires recognition that resources and capacity can vary greatly in organization, and search approaches should be modular to make algorithmic fairness a distributed and accessible practice. Horizontal LDA search as proposed in this paper, along with the relational tradeoff framework, exploits inherent structural differences in model families to surface variable fairness effects, in a light-weight, resource non-intensive manner. This creates a pathway to democratize the practice of fairness, expanding the scope of who can participate in fairness work—not just organizations with deep technical expertise or compute resources, but any institution tasked with making socially consequential decisions. A horizontal LDA search can function independently as a minimum viable fairness approach for resource constrained environments, or in complement preceding a vertical LDA search (i.e., within model optimization and hyperparameter tuning) for organizations that seek a more comprehensive approach. Integrating fairness considerations from the outset of model development enables institutions to better align their decision-making processes with both business objectives and ethical responsibilities, strengthening consumer confidence in AI-driven applications. Overall, advancing the wide-scale adoption of responsible AI practices requires continuous conversation; we encourage fostering ongoing collaboration among industry peers, promoting transparent sector-specific benchmarking, and supporting regulatory guidance, to mitigate harms while maintaining efficiency and effectiveness for AI systems in use.

%\begin{acks}
%We thank our colleagues in NFHA's legal department for their support and review.
%\end{acks}

\bibliographystyle{ACM-Reference-Format}
\bibliography{HMDA_ref}

\appendix

\newpage
 \renewcommand{\thefigure}{A.
 \arabic{figure}}
 \setcounter{figure}{0}

\section{ A Review of Machine Learning Algorithms \label{sec:models}}

This paper primarily focuses on testing five commonly used algorithms for analysis; this includes logistic regression, K-nearest neighbors classification (KNN), classification and regression tree (CART), Gaussian Na\"ive Bayes (NB), and random forest (RF), which are briefly explained below.

\subsubsection{Logistic regression (LR)}

Logistic regression assumes that the probability of the outcome of interest is a linear function of the observed covariates. When the decision boundary, often known as the linear predictor, exceeds zero (equivalently, the predicted probability is greater than 0.5), the predicted outcome is classified as 1, and 0 otherwise. The model is known to be suboptimal in the presence of imbalanced data and with non-linear effects of the covariates. However, it is still preferred over many modern black-box methods due to its transparency. Although there are robust and penalized versions of the model that are optimized for predictions, for this paper, we focus on an ordinary logistic regression model with no regularization.

\subsubsection{K-nearest neighbors classification (KNN)}

The K-nearest neighbors (KNN) algorithm is a non-parametric, supervised machine learning approach utilized for classification and regression tasks. In the KNN algorithm, data points are assigned to a class based on the majority class of their 'K' nearest neighbors in the feature space. The 'K' in KNN signifies the number of neighbors considered for the classification or regression decision. In the context of our focus on classification tasks, the algorithm assigns a class label to a data point through a majority vote from its nearest neighbors. For regression tasks, KNN computes the average value of the K nearest neighbors to predict a continuous outcome. KNN is versatile and does not assume any underlying data distribution, making it suitable for various types of datasets. However, its performance can be sensitive to the choice of the distance metric and the value of K, which may become computationally expensive for large datasets. Additionally, it suffers from the curse of dimensionality if the dimensions of the feature space are large, as there will generally be very few training observations close to the point of interest. For a recent review and in-depth analysis of KNN, we refer interested readers to \citet{syriopoulos2023k}. For this study, we use K=5 as specified in \citet{lee21}. 

\subsubsection{Classification and regression tree (CART)}

Classification and regression trees (CART) introduced by \citet{breiman1984classification} constitutes a nonparametric supervised learning approach that builds tree-like models for both categorical (classification) and continuous (regression) prediction tasks. The algorithm recursively partitions the feature space into distinct regions, assigning a predictive model (either class labels for classification or continuous values for regression) to each resulting segment. CART employs binary splitting at each node, selecting the optimal feature and threshold based on criteria such as Gini impurity for classification or mean squared error for regression. The resulting tree structure provides interpretable decision paths, helping to understand complex relationships within the data. While CART inherently handles non-linear relationships and interactions, it may be prone to overfitting, prompting the adoption of techniques like pruning and ensemble methods for regularization. However, it is an unstable algorithm, particularly when dealing with noisy data or in the presence of outliers.

\subsubsection{Gaussian Na\"ive Bayes (NB)}

Gaussian Na\"ive Bayes (NB) is a probabilistic generative model designed for classification tasks, known for its simplicity and efficiency. It is derived from Bayes' theorem and assumes that features are conditionally independent given the class label, thereby simplifying the computation of probabilities. In the context of continuous features, Gaussian NB models the likelihood of each feature's distribution as Gaussian, facilitating the estimation of class probabilities. This algorithm is particularly well-suited for datasets where feature-independence assumptions hold to a reasonable extent. Despite its inherent simplicity and the naive assumption of feature independence, Gaussian NB often performs remarkably well, especially in scenarios with limited training data. A recent review including software implementation and extensions of the method can be found in \citet{wickramasinghe2021naive}.

\subsubsection{Random Forest (RF)}

Random forest (RF) is an ensemble learning approach for both classification and regression tasks, known for harnessing the collective strength of multiple decision trees to deliver robust and well-generalized predictions \citep{breiman2001random}. It operates by creating bootstrap samples from the training data and randomly selecting features during tree induction. The ensemble's prediction is aggregated through majority voting (classification) or averaging (regression) to reduce variance and improve model accuracy. This ensemble approach effectively addresses overfitting issues common in single decision trees, while maintaining feature importance measures for understanding the impact of individual features on the prediction. However, choosing the optimal number of trees and tuning hyperparameters remain crucial steps in optimizing RF performance in various applications. Although there are newer techniques for pruning RF (\citet{manzali2023random}), we use the hyperparameter specifications from \citet{lee21}.

\section{Recommendations for Operationalizing Fairness Search}\label{policy recommendations}

In hopes of trying to amplify best practices and encourage broader adoption of responsible AI development methodologies, this paper also offers four recommendations based on our findings for public and private sector entities trying to develop their risk management frameworks and LDA searches within the existing AI legal landscape.  
 
\subsection{Recommendation 1: Develop an LDA Forward Development Process }

Any serious movement towards transitioning to automated decision-making systems in the public and private sectors merits utilizing a rigorous, justifiable, and transparent process in the early model development stage. The foundation of this is predicated on a structured and well-documented LDA search. What this looks like might vary based on institutional capacity to establish what is "reasonable" for their particular case. For some, an initial horizontal LDA search, surveying across models for fairness gains, may be sufficient, while others may want to consider more comprehensive approaches as used in vertical LDA searches, e.g., tuning, or some combination of both. Regardless, a search should be undertaken as the current disparate impact legal regime expects that should precautions not be taken or deemed insufficient, an organization could then expect to bear costs associated with re-evaluating, re-developing, and replacing the product or service with an LDA \citep{black2024}. It is preferable then that rather than looking at fairness assessments as an isolated step in the model development cycle, organizations should aim to have interoperable internal systems in place that can responsibly steward and deploy data across the model lifecycle. 

From the initial project stages, prerequisite data used for disparate impact analysis of an algorithm should be handled within appropriate norms of responsible data governance, i.e., ethical souring and limiting what contexts sensitive personal data is used in to only what is necessary. A framework for disparate impact analysis should be selected, such as the methodology used in this study, and the outcomes of the analysis should then be evaluated within the context of broader project priorities. Overall, an institution should be able to clearly articulate which model options were considered, what fairness-accuracy trade-offs existed in choosing to adopt one algorithm over another, and ultimately, why one algorithm was preferred for the scope of the stated task. Documentation of this information and periodic monitoring of disparate impact should be considered an inherent part of an entity's risk management and compliance framework.

\subsection{Recommendation 2: Consider a Developer-first Approach to Documentation and Compliance}

State-based AI legislation is demonstrating an emerging interest in AI accountability as primarily a developer-oriented responsibility \citep{colorado2024}. With model multiplicity being a widely accepted theory in the AI community, there is almost always a less discriminatory algorithm should one look for it; thus the law is increasingly resting the burden of model harm in the development process if efforts to circumvent potential harm are not proven. Against this context, organizations should consider adjusting their risk management and compliance strategies to focus on the role of their engineering team in the product development process. 

The documentation process of an organization's LDA search, complete with clear articulation of the trade-offs considered in adopting one algorithm over another, is a responsibility that falls within the unique upstream purview of developers by virtue of them being architects of the model. By prioritizing firm resources towards this high-impact liability area, in addition to consciously implementing LDAs, organizations avoid the basis of the majority of algorithmic-based harm suits through clearly showing compliance with the existing legal regime and federal guidelines on automated decision-making. It is also not a given that auditors, a compliance team, or other relevant stakeholders will be technically literate, making it important to ensure that documentation is understandable to a variety of audiences, for example, the visualized trade-off diagram included in this paper (Figure \ref{fig_Trade-off}).

\subsection{Recommendation 3: Ensure Adequate Government Support for LDA Search \& Development} 

To ensure that financial institutions conduct robust searches for LDAs, fairness obligations must be accompanied by clear guidance, concrete examples, and best-practice principles \citep{chien2024}. Policymakers should provide comprehensive guidelines in key areas, including appropriate debiasing techniques, recommendations for the proper depth of LDA searches, considerations for LDA viability, and suitable fairness metrics in a variety of different contexts \citep{chien2024}.

While prescriptive standards may be premature given the current evolving nature of this field, regulators should actively promote consensus on best practices and encourage institutions to adopt emerging techniques.  Additionally, explicit outcome expectations should be set regarding the frequency of LDA searches, their integration into model development, and their continuous re-evaluation as external conditions change. Finally, regulators must offer sector-based direction on how to select suitable fairness objectives appropriate for different products and consumer segments. By supporting LDA research and development through structured guidance and regulatory oversight, policymakers can help financial institutions implement fairer decision-making processes and mitigate disparate impact more effectively.

\subsection{Recommendation 4: Industry Peers Should Benchmark Algorithms and Share Best Practices}

Beyond regulatory guidance by policymakers, industry players have a critical role in the process of defining and establishing standards and best practices for considering LDAs. For example, model developers should share the standard acceptable range for model performance could establish a baseline for viable models. Achieving this goal will require coordination among industry coalitions and consortiums to facilitate the sharing of LDA search methodologies, and ensure proper sector-specific benchmarking standards for fairness. Given that fairness measures and disparate impact metrics may vary widely across different industry contexts, it will be essential to maintain transparency in utilized performance metrics, methodologies, and fairness assessments to ensure accountability and foster continuous improvement. Standardized benchmarks will enable financial institutions to compare the effectiveness of various approaches, identify emerging best practices, and refine fairness-focused model development, while retaining proprietary access to their models. 

Each of these recommendations are independent, yet complementary steps for the at-scale adoption of AI accountability in practice. They assist in translating the normative values responsible AI encourages into systems and processes, establishing ubiquitous, standardized AI fairness across the board. As outlined, developers, institutions, and governments each have role-specific responsibilities to develop systemic and structured expectations on the process, documentation, and standards required to achieve this goal. Auditing and compliance work, as a result, can become less opaque, and the use and development of AI systems can be demystified for increased process efficiency and public trust cultivation.

 \section{Data Visualizations}

\newpage
\subsection{HMDA 2021 Home Ownership Race Share for Approved and Denied Loans}
\begin{figure}[htbp]
  \centering
  \includegraphics[width=0.7\textwidth]{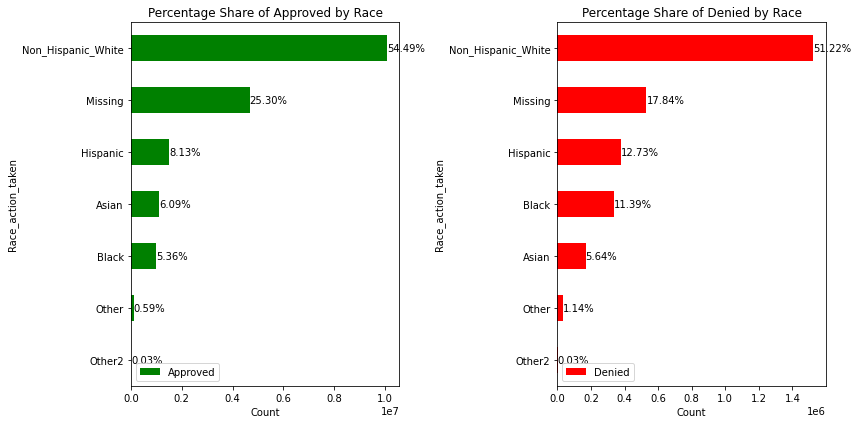}  
   \caption{\hspace{-1em}}
   \label{fig_Race_Loan_ApprovedDenied}
\end{figure}

\subsection{HMDA 2021 Home Ownership Race Share for Average Loan Amounts}

\begin{figure}[htbp]
  \centering
  \includegraphics[width=0.8\textwidth]{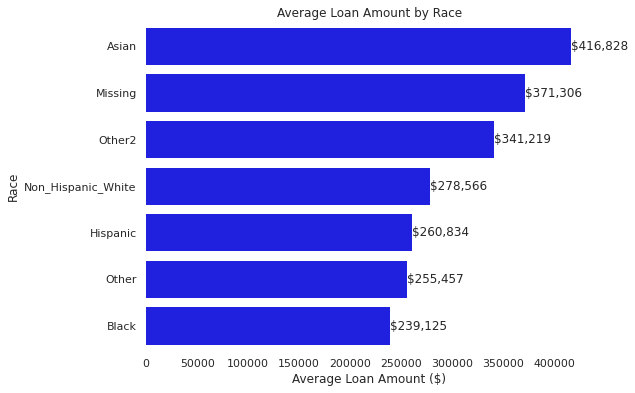}  
 \caption{\hspace{-1em}}
  \label{fig_AvgLoanAmount}
\end{figure}

\subsection{HMDA 2021 Home Ownership Race Share for Average Interest Rates}
\begin{figure}[htbp]
  \centering
  \includegraphics[width=0.9\textwidth]{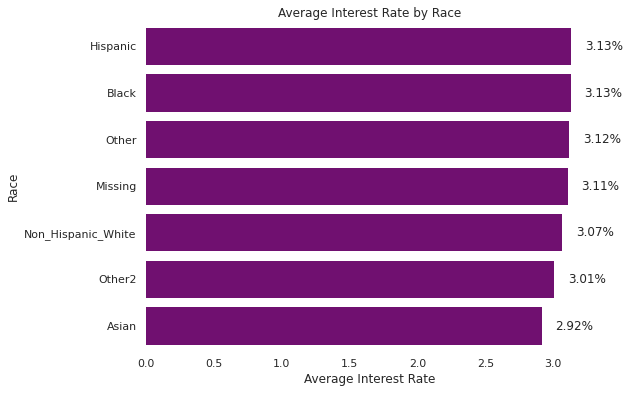}  
   \caption{\hspace{-1em}}
  \label{fig_AvgIntRate}
\end{figure}

\section{Features used in Model Development}

  \begin{table}[htbp]
  \centering
   \scriptsize % Adjust the size as needed
   \label{tab:example3}
   \begin{tabular}{|p{4.9cm}|p{1.4cm}|p{8cm}|}
     \hline
     \multicolumn{1}{|c|}{\textbf{Feature}} & \multicolumn{1}{c|}{\textbf{Type}} & \multicolumn{1}{c|}{\textbf{Values}}\\
     \hline								
 	Income	& Numeric & Amount of the covered loan, or the amount applied for \\	
     \hline
 	Applicant Sex & Categorical	& Male, Female, Unknown, not applicable \\
     \hline
 	Race	& Categorical & 	Black, Non-Hispanic White, Hispanic, Asian, Other2, Other, as outlined in Table 3\\
     \hline	
 	Occupancy Type	&  Categorical & Principal residence, Second residence, Investment property\\
     \hline	
     Derived dwelling category	&  Categorical & Single Family (1-4 Units): Site-Built Multifamily:Site-Built (5+ Units), Single Family (1-4 Units): Manufactured,Multifamily:Manufactured (5+ Units)						\\	
     \hline	
     Loan Purpose	&  Categorical & Home purchase, Home improvement, Refinancing, Cash-out refinancing,Other purpose\\	
     \hline	
     Loan Type	&  Categorical & Conventional (not insured or guaranteed by FHA, VA, RHS, or FSA), Federal Housing Administration insured (FHA),Veterans Affairs guaranteed (VA), USDA Rural Housing Service or Farm Service Agency guaranteed (RHS or FSA)\\
     \hline
    Loan Amount	&  Numeric & The amount of the covered loan, or the amount applied for\\
     \hline
   Tract Population	&  Numeric & Total population in tract						\\	
     \hline
     Tract Minority Population	&  Numeric & Percentage of minority population to total population for tract, rounded to two decimal places\\	
     \hline
     FFIEC msa md median family income	&  Numeric & FFIEC Median family income in dollars for the MSA/MD in which the tract is located (adjusted annually by FFIEC)\\
     \hline
     Tract to msa income percentage	&  Numeric & Percentage of tract median family income compared to MSA/MD median family income	\\
     \hline
     Tract owner occupied units	&  Numeric & Number of dwellings, including individual condominiums, that are lived in by the owner\\
     \hline
     Tract one to four family homes	&  Numeric & Dwellings that are built to houses with fewer than 5 families	\\
     \hline
   \end{tabular}
   \caption{\hspace{-1em}}
   \label{tab_ModelFeatures}
 \end{table}

\end{document}